\pdfoutput=1

\documentclass[11pt]{article}
\usepackage{float}

\usepackage[]{acl2023}

\usepackage{xspace}
\usepackage[utf8]{inputenc}
\usepackage{pgfplots}
\usepackage{dsfont}
\DeclareUnicodeCharacter{2212}{−}
\usepgfplotslibrary{groupplots,dateplot}
\usetikzlibrary{patterns,shapes.arrows}
\pgfplotsset{compat=newest}

\usepackage{tikzscale}
\usepackage{amsmath}
\newcommand{\probP}{\text{I\kern-0.15em P}}
\newcommand{\KARL}{$\text{KAR}^\text{3}\text{L}$\xspace}
\newcommand{\KARLDelta}{$\text{KAR}^\text{3}\text{L}+\Delta$\xspace}

\usepackage{multirow, colortbl}

\usepackage{tabularx,booktabs}
\usepackage{makecell}
\usepackage{mathtools}

\usepackage[normalem]{ulem}
\useunder{\uline}{\ul}{}

\usepackage{graphicx}

\usepackage{booktabs}
\usepackage[normalem]{ulem}
\useunder{\uline}{\ul}{}

\newcommand{\specialcellleft}[2][l]{%
\begin{tabular}[#1]{@{}l@{}}#2\end{tabular}}

\usepackage{times}
\usepackage{latexsym}
\usepackage{adjustbox}

\usepackage[T1]{fontenc}
\usepackage{amsmath}
\usepackage{amssymb}
\usepackage{booktabs}
\usepackage{multirow}
\usepackage{scalerel,xparse}
\usepackage[utf8]{inputenc}

\usepackage{cleveref}
\usepackage{microtype}
\usepackage{enumitem}
\crefformat{section}{\S#2#1#3}
\crefformat{subsection}{\S#2#1#3}
\crefformat{subsubsection}{\S#2#1#3}

\definecolor{UMDred}{HTML}{ed1c24}
\definecolor{UMDyellow}{HTML}{f5a020}
\definecolor{CustomGreen}{HTML}{1FC801}
\definecolor{CustomBlue}{HTML}{324aa8}

\usepackage[a-1b]{pdfx}

\usepackage{framed}
\usepackage{mdwlist}
\usepackage{siunitx}
\usepackage{latexsym}
\usepackage{colortbl}
\usepackage{xcolor}
\usepackage{nicefrac}
\usepackage{booktabs}
\usepackage{fnpct}
\usepackage{amsfonts}
\usepackage[T1]{fontenc}
\usepackage{bold-extra}
\usepackage{amsmath}
\usepackage{amssymb}
\usepackage{bm}
\usepackage{graphicx}
\usepackage{mathtools}
\usepackage{microtype}
\usepackage{multirow}
\usepackage{multicol}
\usepackage{xpatch}
\usepackage{latexsym,comment}
\usepackage[normalem]{ulem}

\newcommand*{\missingreference}{{\Huge \colorbox{red}{?reference?}}}
\newcommand*{\missingcitation}{{\Huge \colorbox{red}{?citation?}}}

\makeatletter
\xpatchcmd{\@setref}{\bfseries}{\missingreference}{}{}
\def\@citex[#1]#2{\leavevmode
    \let\@citea\@empty
    \@cite{\@for\@citeb:=#2\do
        {\@citea\def\@citea{,\penalty\@m\ }%
            \edef\@citeb{\expandafter\@firstofone\@citeb\@empty}%
            \if@filesw\immediate\write\@auxout{\string\citation{\@citeb}}\fi
            \@ifundefined{b@\@citeb}{\hbox{\reset@font\missingcitation}%
                \G@refundefinedtrue
                \@latex@warning
                {Citation `\@citeb' on page \thepage \space undefined}}%
            {\@cite@ofmt{\csname b@\@citeb\endcsname}}}}{#1}}
\makeatother

\newcommand{\gem}[1]{\mbox{\textsc{gem}}}
\newcommand{\abr}[1]{\textsc{#1}}

\newcommand{\g}{\, | \,}

\newcommand{\hidetext}[1]{}
\newcommand{\ignore}[1]{}

\newcommand{\smallurl}[1]{ \begin{tiny}\url{#1}\end{tiny}}

\definecolor{lightblue}{HTML}{3cc7ea}
\definecolor{CUgold}{HTML}{CFB87C}
\definecolor{grey}{rgb}{0.95,0.95,0.95}
\definecolor{ceil}{rgb}{0.57, 0.63, 0.81}
\definecolor{UMDred}{HTML}{ed1c24}
\definecolor{UMDyellow}{HTML}{ffc20e}

\title{\KARL: Knowledge-Aware Retrieval and Representations aid Retention and Learning in Students}

\author{Matthew Shu$^{1*}$ \hspace{0.5cm} Nishant Balepur$^{2*}$  \hspace{0.5cm} \textbf{Shi Feng}$^{3*}$ \hspace{0.5cm}  \textbf{Jordan Boyd-Graber}$^{2}$\\
  $^{1}$Yale University \hspace{0.5cm}
  $^{2}$University of Maryland \hspace{0.5cm}
  $^{3}$George Washington University \hspace{0.5cm} \\
  \texttt{matthewmshu@gmail.com}, \hspace{0.2cm} \texttt{nbalepur@umd.edu}
}

\begin{document}
\maketitle
\renewcommand{\thefootnote}{\fnsymbol{footnote}}
\footnotetext[1]{Equal contribution.}
\renewcommand{\thefootnote}{\arabic{footnote}}

\begin{abstract} {

Flashcard schedulers rely on 1) \textit{student models} to predict the flashcards a student knows; and 2) \textit{teaching policies} to pick which cards to show next via these predictions.
Prior student models, however, just use study data like the student's past responses, ignoring the text on cards. 
We propose \textbf{content-aware scheduling}, the first schedulers exploiting flashcard content.
To give the first evidence that such schedulers enhance student learning, we build \KARL, a simple but effective content-aware student model employing deep knowledge tracing (DKT), retrieval, and BERT to predict student recall.
We train \KARL by collecting a new dataset of 123,143 study logs on diverse trivia questions.
\KARL bests existing student models in AUC and calibration error.
To ensure our improved predictions lead to better student learning, we create a novel delta-based teaching policy to deploy \KARL online.
Based on 32 study paths from 27 users, \KARL improves testing throughput over SOTA, showing \KARL's strength and encouraging researchers to look beyond~historical study data to fully capture student abilities.
\footnote{Our data and code are available at \url{https://github.com/Pinafore/fact-repetition}.}

}
\end{abstract}

\begin{figure*}[t]
\centering
\includegraphics[width=\linewidth]{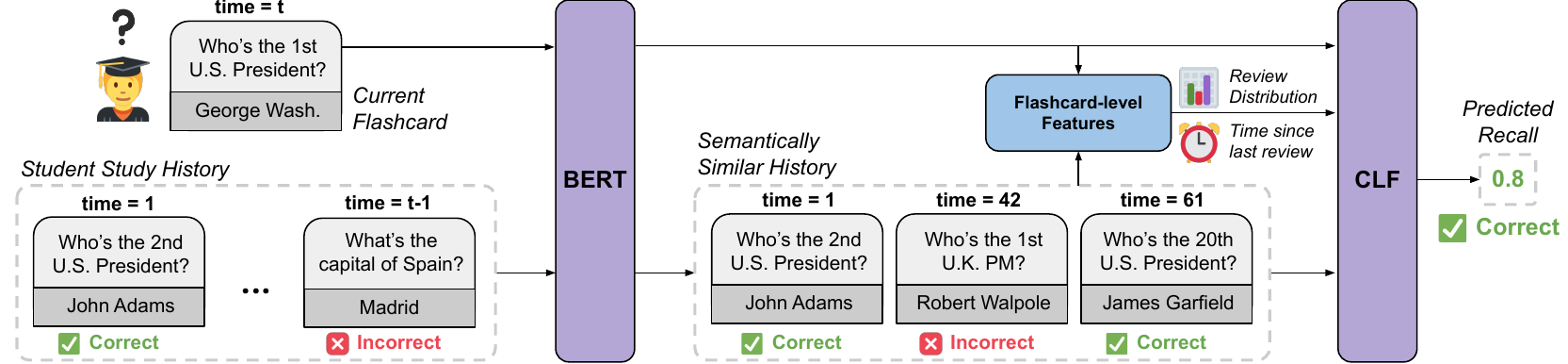}
\caption{\label{fig:model}
Overview of \KARL. Given a current flashcard and the student's study history as inputs, \KARL first uses a BERT retriever to obtain the most semantically similar cards from the study history. Next, the BERT embeddings of these retrieved flashcards, the embedding of the current flashcard, and flashcard-level features (e.g. time since last review), are fed through a classifier (CLF) to predict if the student knows the answer to the current flashcard.
}
\end{figure*}

\section{Introduction}

Flashcards help students learn answers to questions across subjects like trivia, vocabulary, and law \cite{wissman2012and}.
In education, a flashcard \textbf{scheduler} dictates when students review old flashcards and when they learn new ones~\cite{reddy2016unbounded}.
Many schedulers use \textbf{student models} to predict the probability a student can recall a flashcard~\cite{mozer2019artificial}.
A \textbf{teaching policy} then chooses the flashcards to show next based on recall predictions~\cite{reddy2017accelerating}.
Teaching policies ensure personalized learning experiences by understanding a student's learning progress.
Thus, every student model must use informative features to capture a student's knowledge state~\cite{brusilovsky2015open, chrysafiadi2015student}.

To predict recall, existing student models \cite{settles2016trainable, tabibian2019enhancing} just use data from the student's study history, like their past answers and time since last review. 
This data models student behavior, but it ignores a key aspect of the card: its textual content.
Modeling the~relations across flashcard content can enable student models to predict recall even on cards with no study data. 
For instance, if a student studies the question ``Who was the \emph{first} U.S. president'' for the first time, existing student models cannot discriminatively predict if the student knows the answer, as the card lacks study data.
However, if the student already studied ``Who was the \emph{second} U.S. president'' and always answered it correctly, we can infer that the student knows U.S. presidents and likely can recall the first U.S. president.
Existing schedulers cannot make these semantic inferences, limiting their ability to predict and schedule flashcards with no study data. 


We propose \textbf{content-aware flashcard schedulers}---the first schedulers that exploit flashcard content.
We connect this paradigm to Deep Knowledge Tracing \cite[DKT]{piech2015deep}, which embeds student study histories~\cite{shin2021saint+}~to predict study correctness.
Some DKT models, like LM-KT \cite{srivastava-goodman-2021-question}, enhance these embeddings with language models, providing a strong basis for semantic inferences.
While DKT models show promise on offline benchmarks, their adoption in real-world flashcard scheduling systems is absent, as there lacks concrete evidence that such models can improve learning (\cref{section:model_motivation}).
Thus, our goal is to design a baseline content-aware scheduler using DKT and be the first to show that this model can successfully enhance student learning, motivating the benefits and potential of the paradigm.

Towards content-aware scheduling, we develop \textbf{\KARL}, a DKT student model using \textbf{K}nowledge-\textbf{A}ware \textbf{R}etrieval and \textbf{R}epresentations for \textbf{R}etention and \textbf{L}earning (Figure~\ref{fig:model}).
While DKT models usually embed the student's full study history, \KARL uses a BERT~retriever \cite{lee-etal-2019-latent} to find a subset of semantically similar cards in the study history, making \KARL the first retrieval-based student model (\cref{subsection:flashcard_retrieval}).
Retrieval yields just study history with similar content to the current card, allowing \KARL to jointly improve efficiency and omit noisy parts of the history that distract the representation of the student's knowledge (\cref{subsection:ablation}). \KARL then encodes the retrieved history and current flashcard with the student's study data and BERT. This lets \KARL accurately predict recall even on unseen cards, as \KARL can infer student knowledge via study data from semantically related~flashcards.

A general student model like \KARL cannot be trained on existing datasets, as they do not release flashcard text or have narrow domains~\cite{settles2016trainable, selent2016assistments}.
Thus, we curate flashcards using content-rich trivia questions on diverse topics like fine arts, history, and pop culture.
We develop a flashcard app and deploy our cards to 543 learners (\cref{section:dataset})---forming a new dataset of 123,143 study logs to train content-aware models.
Compared to baselines~\cite{settles2016trainable, srivastava-goodman-2021-question, FSRS},
\KARL gives the most accurate and well-calibrated recall predictions, demonstrating the offline strength of content-aware student models (\cref{section:offline_evaluation}).

Researchers typically stop at offline evaluation to claim scheduler superiority, but this does not assess a scheduler's main goal: enhancing student learning.
To ensure the improved predictions in \KARL translate to better learning outcomes for students, we design the first online user study~comparison with FSRS, the SOTA scheduler (\cref{section:online_evaluation}). To do so, we create a novel delta-based teaching policy that picks flashcards predicted to enhance learning after a specified time delta (\cref{subsection:teaching_policy}).
We equip the \KARL student model with this policy to form~\KARLDelta, the first content-aware~scheduler to aid learning.

We have 27 new users study in our app and test their medium-term learning  : \textbf{response accuracy}, the number of new flashcards learned,~and \textbf{response time}, the time taken to recall answers. In 32 six-day studies, \KARLDelta maintains learning accuracy while reducing recall time, showing testing throughput gains over FSRS (\cref{subsection:test_mode_results}). \KARLDelta, a~baseline content-aware scheduler, bests SOTA,~revealing the strength of \KARL, motivating future works to build better content-aware schedulers, and encouraging researchers to look beyond study data to capture student abilities. Our contributions~are:\\
\noindent \textbf{1)} We introduce content-aware scheduling, the first flashcard schedulers that exploit flashcard content.

\noindent \textbf{2)} We implement \KARL, a simple but effective content-aware student model that uses DKT, BERT, and retrieval. \KARL employs a novel delta-based teaching policy to facilitate online scheduling.

\noindent \textbf{3)} We collect and release a dataset of 123,143 study logs from 543 users on diverse trivia flashcards. 

\noindent \textbf{4)} We design an online evaluation on two facets of medium-term learning to prove that content-aware models like \KARL can effectively aid learning.

\section{Related Work}

\paragraph{Flashcard Scheduling:}

Flashcards help students recall answers to questions, ranging from vocabulary \cite{vocab1, vocab2} to medicine \cite{medicine1, medicine2}.
The order and spacing of flashcards when studying strongly affect the student's ability to recall information in the future (e.g. exam time) \cite{kornell2009optimising}, leading to research in systems that optimally schedule flashcards. Early models like Leitner \cite{leitner1974so} and SuperMemo-2 \cite{wozniak1990optimization} use rule-based heuristics to pick review dates. Subsequent schedulers draw~from cognitive theory, deploying teaching policies with student model predictions. \citet{settles2016trainable} design half-life regression (\textsc{HLR}) student models as trained power-law and exponential forgetting curves \cite{ebbinghaus_memory_1913}, theoretical memory decay models derived from empirical research.

Later works extend HLR and build teaching policies to schedule flashcards based on learned parameters and predictions from forgetting curve student models. \citet{reddy2017accelerating} optimize their teaching policy with reinforcement learning, while MEMORIZE \cite{tabibian2019enhancing} and SELECT \cite{upadhyay_large-scale_2020} minimize cost values to optimize review times. Recent models like~FSRS and DHP enhance these methods by optimizing spaced repetition via stochastic shortest path algorithms and time series data \cite{FSRS, su2023optimizing}.
 


Notably, the student models in these schedulers cannot discriminate between unseen flashcards and must arbitrarily show new cards, such as ordering by cards' creation dates \cite{elmes_anki_2021}. 
Content-aware scheduling---predicting student recall using the content on cards---addresses this limitation.

\paragraph{Deep Knowledge Tracing:}

\KARL uses deep knowledge tracing (\textsc{DKT}) \cite{shin2021saint+, KTSurvey}: neural models that predict if a student has knowledge of a specified concept, subject, or study item (e.g. flashcard).
The first DKT model used an RNN to capture the temporal dynamics of a student's study history \cite{piech2015deep}. Later works refine this model by incorporating graph representations \cite{graph2, graph1}, forgetting features \cite{Chen2017TrackingKP, forget2}, and memory structures \cite{memory1, memory2}.

Recent DKT models embed question content to improve study history encodings \cite{su2018exercise, yin2019quesnet, lee-etal-2024-difficulty}. However, these models cannot be directly adapted for scheduling. LM-KT \cite{srivastava-goodman-2021-question} is based on GPT-2 so it can embed diverse questions, but its causal language modeling objective impedes the use of flashcard features, as next-token prediction models struggle to reason over numerical inputs \cite{mcleish2024transformers}. Other text-aware models can use flashcard features but are designed for the math domain \cite{liu2019ekt} and need content annotations of study items to discern relevant items in the study history. Conversely, \KARL is a classifier that can better encode numerical inputs via a feature embedding layer, and uses retrieval which can find relevant content without content annotations, combining the strengths of existing DKT models.

DKT models show promise when trained on offline benchmarks that assess if models can predict student study correctness, often the top-performing models \cite{KTSurvey}. 
However, there is no concrete evidence that DKT models can or should be adopted to facilitate student learning in online applications like flashcard learning software. 
We bridge this gap by solving practical issues of DKT models to motivate their adoption, via: 1) the first retrieval-augmented DKT model to mitigate inefficiency and study history noise; 2) a new dataset of study logs on diverse questions to train content-aware models; and 3) a delta-based teaching policy and user study to prove DKT can enhance learning.

\paragraph{NLP in Education:}

Flashcard scheduling is just one educational task that benefits from \textsc{NLP} \cite{litman2016natural}.
Recent research in this area includes writing education content \cite{cui-sachan-2023-adaptive}, designing educational chatbots \cite{tyen-etal-2022-towards, liang-etal-2023-chatback, siyan2024eden}, exploring test-taking strategies like process of elimination \cite{ma2023poe, balepur-etal-2024-easy}, understanding student misunderstandings \cite{wang-etal-2024-backtracing}, and creating mnemonic devices \cite{lee2023smartphone}.
Our contributions, such as the introduction of content-aware scheduling and release of a new diverse study history dataset, will facilitate further research in NLP-powered educational tools.

\section{Where Are Content-Aware Schedulers?} \label{section:model_motivation}

Our core method for content-aware scheduling (\cref{section:model_design}) uses BERT, DKT, and retrieval---techniques with proven benefits---so why have they not yet been adopted for scheduling?
Our work identifies and addresses three key criteria needed to inspire broader adoption:
1) an effective DKT student~model that does not need content annotations on study items and can efficiently be deployed (\cref{section:model_design});
2) a large, diverse dataset to train content-aware models (\cref{section:dataset});
and 3) a user study to confirm content-aware schedulers benefit learning (\cref{section:online_evaluation}).
As a whole, these challenges indicate that the absence of content-aware schedulers stems from a lack of evidence showing that they meaningfully improve student learning.

We reveal that with the right~modeling choices (\cref{section:model_design}), content-rich datasets (\cref{section:dataset}), and thorough user studies (\cref{section:online_evaluation}), we can adeptly combine the strengths of language models, DKT, and retrieval to show that content-aware schedulers aid student learning.

\section{\KARL Student Model Design} \label{section:model_design}


Our student model \KARL builds on Deep Knowledge Tracing \cite[DKT]{naeini2015obtaining} and uses as inputs: 1) the flashcard $f_{t}$ shown to the student at time $t$; and 2) a history of all past flashcards studied by the student $\mathcal{F} = \{f_1, f_2, ..., f_{t-1}\}$.
We also assume a flashcard $f$ can be mapped to study data $\mathcal{X}(f)$, like the student's total correct responses to $f$ and its time since last review. 
Using these inputs, \KARL predicts correctness~$a_t \in \{0, 1\}$ denoting whether the student will answer card $f_t$ correctly. 

\KARL predicts $a_t$ in two steps (Figure \ref{fig:model}). First, \KARL uses a BERT retriever $p(f_t \g f_i)$ to find the top-$k$ flashcards $\mathcal{F}' \subseteq \mathcal{F}$ most semantically relevant to $f_t$. Next, \KARL feeds $f_t$ and $\mathcal{F}'$ as inputs to a classifier $p(a_t \g f, \mathcal{F}')$, which represents the current flashcard $f$ and each retrieved flashcard in $\mathcal{F}'$ with BERT embeddings and the features from $\mathcal{X}(z)$. We describe both of these steps~next.

\subsection{Flashcard Retrieval} \label{subsection:flashcard_retrieval}

A student's study history $\mathcal{F}$ is often long and diverse.
DKT models encode all of $\mathcal{F}$, degrading efficiency, and their $f_t$ predictions may also be worse if they use study data in $\mathcal{F}$ that is unlike $f_t$~(\cref{subsection:ablation}).
For example, if a student studies both math and history, embedding all of $\mathcal{F}$ may worsen predictions on history cards, as the study data on math is irrelevant.
DKT datasets have predefined subject or knowledge component labels on study items to help models discern relevance \cite{koedinger2012knowledge}. However, flashcard apps support user-created cards that may not fall into these categories, so we assume no access to study item labels.

To solve these issues, we design the first retrieval-augmented student model.
In generation, retrievers limit noise in large corpora and improve efficiency by picking a subset of relevant items \cite{lewis2020retrieval, balepur2023expository}.
We retrieve the flashcards $\mathcal{F}'$ from the student's study history $\mathcal{F}$ with the most similar semantic representations to the~current flashcard $f_t$. 
This ensures \KARL makes predictions using the study history most similar to the card $f_t$, reducing the total cards to embed and study history noise (\cref{subsection:ablation}) without study item annotations.
Further, \cref{subsection:qualitative_analysis2} shows that retrieval has the additional benefit of uncovering concepts in flashcards.

We obtain the semantic similarity between each study history card $f_i \in \mathcal{F}$ and the current flashcard $f_t$ via the dot product of $f_i$ and $f_t$, represented by pretrained BERT \cite{devlin-etal-2019-bert} embeddings:
\begin{align} 
\textbf{d}(f_i) & = \text{BERT}(f_i), \\ 
\textbf{q}(f_t) & = \text{BERT}(f_t), \\
\label{eq:retriever}p(f_t \g f_i) & \sim \textbf{d}(f_i)^T \textbf{q}(f_t).
\end{align}
\noindent Maximum Inner-Product Search \cite{shrivastava2014asymmetric} finds the $k$-highest values for $p(f_t \g f_i)$, forming the top-$k$ relevant cards $\mathcal{F}' \subseteq \mathcal{F}$ to $f_t$. 

\subsection{Feature Representation} \label{subsection:feature_representation}

After finding the semantically relevant flashcards $\mathcal{F}'$, we represent the flashcards in the history $f_i \in \mathcal{F}'$ and the current flashcard $f_t$ with features for predicting student recall. Along with BERT embeddings from \cref{subsection:flashcard_retrieval}, we also use the flashcard-level heuristics from prior student models \cite{settles2016trainable, tabibian2019enhancing}. We define $\mathcal{X}(f)$ as a set of features that represent the study data on flashcard $f$, such as the distribution of the student's past responses to $f$ and the time since its last review. We detail all features in Appendix~\ref{appendix:subsection:classifier_features}.

Using BERT embeddings and study data as features, we train a classifier $p(a_t \g f_t, \mathcal{F}')$ that uses the current card $f_t$ and retrieved history $\mathcal{F}'$ to predict correctness $a_t$, which is 0/1 if the student recalls $f_t$~incorrectly/correctly. We feed study data $\mathcal{X}(f)$ and BERT embeddings for study history flashcards $f_i \in \mathcal{F}'$ and the current flashcard $f_t$ to a linear layer to compute a hidden state $h \in \mathbb{R}^{768}$. We then feed $h$ to a final linear layer to predict $a_t$. We train $p(a_t \g f_t, \mathcal{F}')$ to minimize the cross-entropy loss $\lambda$ of predicting $a_t$, using input representations of cards $f$ and $\mathcal{F}'$, via mini-batch gradient descent.


\section{Training \KARL}
\label{section:dataset}

We now train the \KARL student model.
We describe our data collection platform (\cref{subsection:data_collection_platform}), create flashcards (\cref{subsection:flashcard_creation}), collect student study data (\cref{subsection:collecting_student_data}), and outline our training procedure (\cref{subsection:training_setup}).


\subsection{Data Collection Platform} \label{subsection:data_collection_platform}

\begin{figure}[t]
\centering
\fbox{\includegraphics[width=0.96\linewidth]{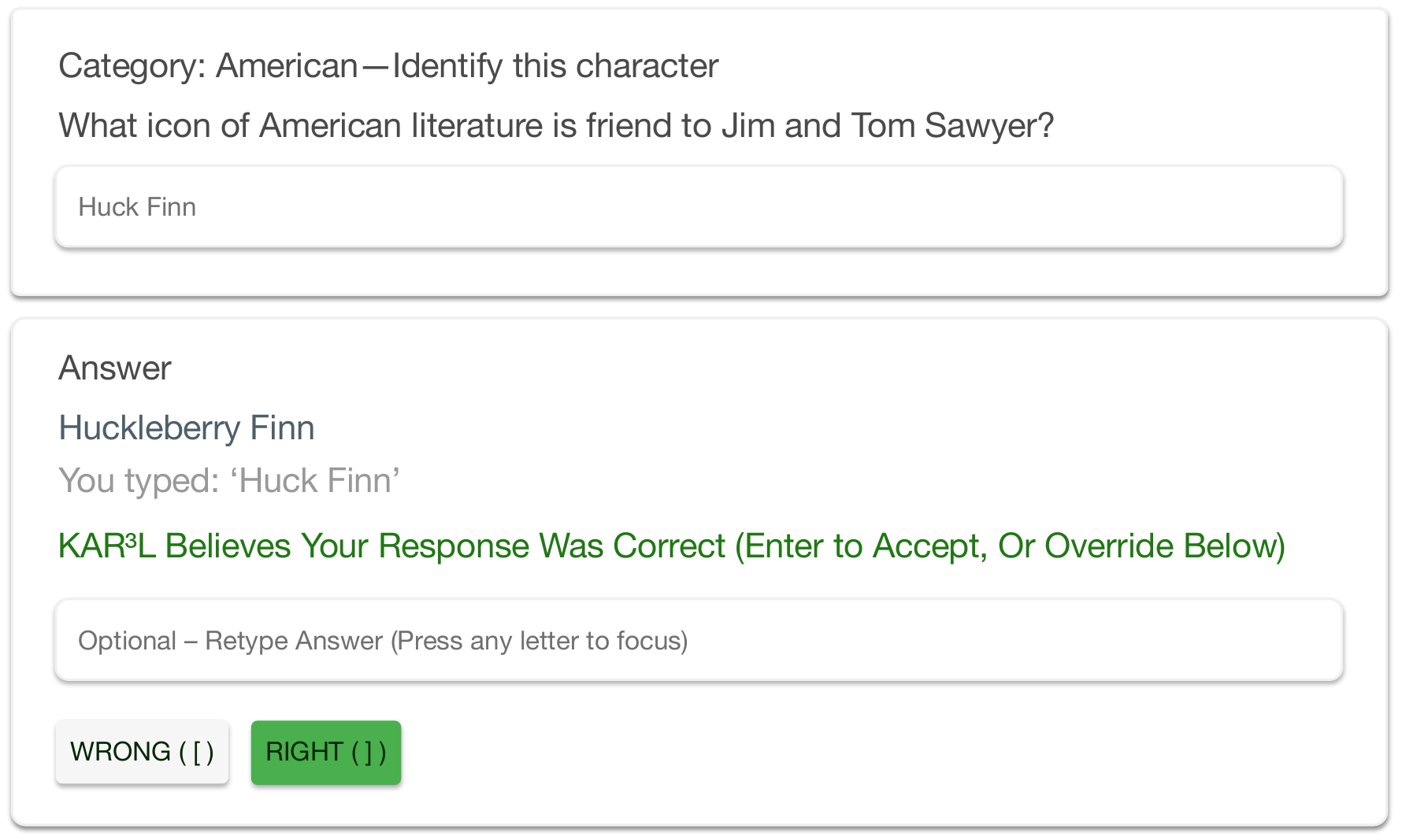}}
\caption{\label{fig:ui}
Screenshot from our web-based flashcard app after a user submits their answer to a literature flashcard.
}
\end{figure}

If we want \KARL to cater to users studying varied topics, we must train on flashcards with diverse content.
However, existing KT and flashcard datasets are domain-specific and include no flashcard text or just a single vocab word, making them unfit for our study. ASSISTments \cite{selent2016assistments}, the most widely-used KT dataset~\cite{KTSurvey}, is restricted to arithmetic; Duolingo's spaced repetition dataset \cite{settles2016trainable} and EdNet~\cite{choi2020ednet} focus on English language learning. 
Hence, to assess how \KARL captures semantic ties across diverse topics, we build our own platform to collect data from real learners, based on a web and mobile flashcard app (Figure~\ref{fig:ui}). 
\subsection{Flashcard Creation} \label{subsection:flashcard_creation}

With diversity in mind, we turn to trivia questions from the QANTA dataset~\cite{Rodriguez2019QuizbowlTC} to make flashcards. QANTA questions are multi-sentence, where each subsequent sentence in the question points to the same answer with decreasing difficulty. The questions span eleven topics, including literature, history, fine arts, pop culture, and mythology. To create cards from these questions, we sample a subset of the dataset and use a sentence from the question as the front of the flashcard and its answer as the back of the card. In total, we curate 23,918 unique flashcards spanning 11 diverse topics (details and examples in Appendix~\ref{appendix:subsection:flashcard_examples}).



\subsection{Study Data Collection} \label{subsection:collecting_student_data}

We recruit users to study cards (\cref{subsection:flashcard_creation}) in~our app from English trivia forums.
Users study with three schedulers; 
two are Leitner~\cite{leitner1974so} and SM-2~\cite{wozniak1990optimization}, popular heuristic schedulers. 
The third is a DKT model~(Appendix~\ref{appendix:cold_start}) trained on the Protobowl dataset~\cite{boyd2012besting}. 
Over four months, 543 users gave 123,143 study logs. 
Each log has the card $f_t$ studied by the user at time $t$, past study data $\mathcal{X}(f_t)$ on $f_t$, and label $a_t$ denoting if the student answered $f_t$ correctly. 
Tables~\ref{appendix:table:data_qual} and \ref{appendix:table:data_quant} show all dataset columns.
Users are referenced by ID and we award \$200 to the fifteen users who study the most flashcards.


\begin{table*}[t]
\small
\centering
\renewcommand{\arraystretch}{0.8}
\setlength{\tabcolsep}{3.9pt}
\begin{tabular}{@{}lcccccccc@{}}
 & \multicolumn{4}{c}{\textit{Seen Cards}} & \multicolumn{4}{c}{\textit{Unseen Cards}} \\ \toprule
\multicolumn{1}{l|}{Model} & AUC ($\uparrow$) & ECE ($\downarrow$) & Acc Correct ($\uparrow$) & \multicolumn{1}{c|}{Acc Incorrect ($\uparrow$)} & AUC ($\uparrow$) & ECE ($\downarrow$) & Acc Correct ($\uparrow$) & Acc
Incorrect ($\uparrow$)  \\ \midrule
\multicolumn{1}{l|}{HLR}  & 0.370 & 0.387 & 0.738 & \multicolumn{1}{c|}{0.144} & - & - & - & - \\
\multicolumn{1}{l|}{Leitner} & 0.752 & 0.234 & 0.833 & \multicolumn{1}{c|}{0.380}  & - & - & - & - \\
\multicolumn{1}{l|}{SM-2} & 0.660 & 0.200 & 0.826 & \multicolumn{1}{c|}{0.419} & - & - & - & - \\
\multicolumn{1}{l|}{FSRS} & 0.752 & 0.111 & 0.921 & \multicolumn{1}{c|}{\textbf{0.524}} & - & - & - & - \\
\multicolumn{1}{l|}{LM-KT} & 0.658 & 0.285 & 0.805 & \multicolumn{1}{c|}{0.327}  & 0.684 & 0.234 & 0.667 & 0.590 \\
\multicolumn{1}{l|}{GPT-3.5} & 0.427 & 0.544 & 0.613 & \multicolumn{1}{c|}{0.241}  & 0.519 & 0.481 & 0.467 & 0.571 \\
\multicolumn{1}{l|}{\KARL} & \textbf{0.864} & \textbf{0.091} & \textbf{0.980} & \multicolumn{1}{c|}{0.250}  & \textbf{0.786} & \textbf{0.085} & \textbf{0.680} & \textbf{0.740} \\
\bottomrule
\end{tabular}
\caption{Student model ability to predict which flashcards students know (Accuracy when Correct/Incorrect, AUC) and how well they know them (ECE). $\uparrow$ (and $\downarrow$) denote if higher (or lower) scores are better. Best scores are in \textbf{bold}. \KARL outperforms baselines in 7/8 metrics, showcasing the offline strength of content-aware scheduling.}
\label{table:quant_results}
\end{table*}

\subsection{Training Setup} \label{subsection:training_setup}

We sort study records chronologically and use a 75/25 train/evaluation split for \KARL. We retrieve $k=5$ items from the student's study history for all experiments using FAISS~\cite{johnson2019billion}. Tables~\ref{appendix:table:data_qual}, \ref{appendix:table:data_quant}, and \ref{appendix:table:data_time} list all study data features collected.
We train \KARL using all features for ablations (\cref{subsection:ablation}) and the online evaluation (\cref{section:online_evaluation}), and use just a subset of all features for the other offline experiments (\cref{section:offline_evaluation}), detailed in Appendix~\ref{appendix:subsection:training}.


\section{Offline Evaluation} \label{section:offline_evaluation}

We evaluate \KARL on our dataset to highlight the offline strength of content-aware \textit{student models}.
This result motivates our design of a content-aware \textit{scheduler} (\cref{section:online_evaluation}), eventually allowing us to prove that such schedulers improve student learning~(\cref{subsection:test_mode_results}).

\subsection{Baselines}

We compare \KARL with popular student models: \\
\textbf{1) Half Life Regression (HLR)} models $a_t$ with an exponential forgetting curve \cite{settles2016trainable}. This curve is fit using the student's past study responses and time since the last study of $f_{t}$.\\
\textbf{2) Leitner} moves a flashcard $f_{t}$ up or down numbered slots based on student responses to $f$ \cite{leitner1974so}. We use five slots and calculate $a_t$ as the slot position of $f_{t}$ divided by the total number of slots. \\
\textbf{3) Super Memo (SM)-2} predicts $a_t$ like Leitner does, adjusting its value based on the student's proportion of successful recalls of $f_t$ \cite{wozniak1990optimization}. \\
\textbf{4) \textsc{FSRS}} computes intermediate difficulty, stability, and retrievability scores to schedule flashcards \cite{FSRS}. We take the retrievability score, defined as recall probability, as a prediction for $a_t$.  \\
\textbf{5) \textsc{LM-KT}} is a representative \textsc{DKT} model using language models (GPT-2 Med), like \KARL \cite{srivastava-goodman-2021-question}. Via causal language modeling, it predicts $a_t$  with the input sequence $\mathcal{F}$.  \\
\textbf{5) \textsc{GPT-3.5}} \cite{ouyang2022training} is five-shot prompted to predict $a_t$ based on the five most recent study items in the student's study history $\mathcal{F}$. We use \texttt{gpt-3.5-turbo} and 0 temperature.

\subsection{Evaluation Metrics}


A student model's training objective is to give binary predictions for whether a student can recall a flashcard, but a student's familiarity with flashcards is not binary but on a continuous spectrum.
Thus,~a strong student model must be able to discern cards that are familiar or unfamiliar to students across a range of thresholds, rather than one binary prediction.
We employ two metrics to capture this nuance: area under the \textsc{ROC} curve (\textbf{\textsc{AUC}}) and expected calibration error \cite[\textbf{\textsc{ECE}}]{naeini2015obtaining}; the former measures how well the model predicts across all thresholds, and the latter measures if predictions are well-calibrated.
We also show accuracy when the student answered correctly ($a_t = 1$) and incorrectly ($a_t = 0$)---the models' training objective.


One benefit of content-aware models is that they can make semantic inferences and predict recall on cards without study data.
To test this for \KARL, LM-KT, and GPT-3.5, we group our metrics by if the student has seen the current flashcard $f_t$ or not. 


\subsection{Quantitative Comparison} \label{subsection:quantitative_comparison}

We study \KARL's ability to model student recall (Table~\ref{table:quant_results}).
On seen cards, \KARL surpasses models in all metrics except for accuracy on incorrect flashcards.
In fact, all models struggle to predict when a student will \textit{incorrectly} answer a flashcard they have already studied, with the best model (FSRS) barely exceeding random guessing (0.524).
Thus, there is a clear opportunity to close this accuracy gap, which could be achieved by focusing on temporal dynamics of student memory \cite{FSRS}.
On unseen cards, \KARL bests LM-KT and GPT-3.5, the only other models able to predict recall on unseen cards, in all metrics.
Further, for \KARL, there is an AUC gap between unseen and seen cards.
Thus, student modeling on cards with no study data is still a challenge and may benefit from even larger LLMs to capture semantics (Appendix~\ref{appendix:ablation_llama}).

\KARL underperforms in Acc Incorrect on seen cards, but we argue AUC and ECE are better indicators of a student model's abilities.
Accuracy uses one cutoff (0.5) to decide if a student can recall a flashcard, but other cutoffs like 0.71 \cite{pavlik2020mobile} and 0.94 \cite{eglingtonOptimizingPracticeScheduling2020} are also valid.
Thus, a robust student model must cater to varied cutoffs, an ability better assessed by AUC and ECE.
Overall, \KARL is the strongest student model, with the best ability to discern which flashcards students know and how well they know them.




\subsection{Ablation Study} \label{subsection:ablation}

To attribute the performance gains in \KARL, we ablate its components.
Adding BERT embeddings and flashcard-level study data both improve ECE and AUC (Table \ref{table:ablation}), showing that content and historical study data both enhance student modeling. 
Further, on unseen cards, where capturing semantic relations should be most useful, No BERT has the worst AUC and ECE. 
Thus, making semantic inferences across flashcards is a valuable strategy for student modeling on unseen flashcards.
\begin{table}[t]
\small
\centering
\setlength{\tabcolsep}{5pt}
\renewcommand{\arraystretch}{0.8}
\begin{tabular}{@{}lcccc@{}}
 & \multicolumn{2}{c}{\textit{Seen Cards}} & \multicolumn{2}{c}{\textit{Unseen Cards}} \\ \toprule
\multicolumn{1}{l|}{Model} & AUC ($\uparrow$) & \multicolumn{1}{c|}{ECE ($\downarrow$)} & AUC ($\uparrow$) & ECE ($\downarrow$) \\ \midrule
\multicolumn{1}{l|}{\KARL BERT} & \textbf{0.780} & \multicolumn{1}{c|}{\textbf{0.108}} & \textbf{0.740} & \textbf{0.124} \\
\multicolumn{1}{l|}{No BERT} & 0.692 & \multicolumn{1}{c|}{0.127} & 0.612 & 0.205 \\
\multicolumn{1}{l|}{No $\mathcal{X}(z)$} & 0.680 & \multicolumn{1}{c|}{0.135} & 0.620 & 0.191 \\ \bottomrule
\end{tabular}
\vspace{-1ex}
\caption{\KARL versus ablations that discard BERT embeddings and discard flashcard-level features $\mathcal{X}(z)$. Both are useful for accurate and calibrated predictions.}
\label{table:ablation}
\end{table}

\begin{table}[]
\small
\centering
\setlength{\tabcolsep}{4pt}
\renewcommand{\arraystretch}{0.8}
\begin{tabular}{@{}lccccc@{}}
\toprule
Metric       & $k=0$ & $k=5$ & $k=10$ & $k=15$ & $k=20$ \\ \midrule
AUC (Seen)   & \textbf{0.864} & \textbf{0.864} & 0.861  & 0.851  & 0.851  \\
ECE (Seen)   & 0.098 & \textbf{0.091} & \textbf{0.091}  & 0.106  & 0.105  \\ \midrule
AUC (Unseen) & 0.776 & \textbf{0.786} & 0.777  & 0.757  & 0.768  \\ 
ECE (Unseen) & 0.111 & \textbf{0.085} & 0.086  & 0.171  & 0.155 \\ \bottomrule
\end{tabular}
\vspace{-1ex}
\caption{Top-$k$ AUC/ECE for new \KARL training runs (AUC same trend). $k=20$ is max without~OOM. Retrieval improves AUC and ECE on unseen cards, but retrieving too many cards can distract recall predictions.}
\label{table:ablation22}
\end{table}
We also assess how retrieval affects \KARL (Table~\ref{table:ablation22}).
On seen cards, \KARL has similar ECE~and AUC with ($k>0$) and without ($k=0$) retrieval, meaning that when study data exists, using just the current flashcard is sufficient to predict recall.
On unseen cards, where study data is absent, retrieval boosts metrics from $k=0$ to $k=5$ and $k=10$.
However, retrieving too many flashcards ($k \geq 15$) on seen and unseen cards harms all metrics versus $k=0$, likely because the excess flashcards are irrelevant, distracting \KARL.
Thus, retrieval helps content-aware schedulers focus on short, relevant contexts for improved recall predictions without needing human annotations to represent relevance.

\subsection{Forgetting Curve Analysis} \label{subsection:qualitative_analysis}

Forgetting curves describe how a student's familiarity with a flashcard changes over time, measured through predicted recall probability \cite{ebbinghaus_memory_1913, murre2015replication}. While many student models explicitly fit to exponential or power-law forgetting curves \cite{upadhyay_large-scale_2020, settles2016trainable}, \KARL does not in exchange for more flexible memory representations. To simulate a forgetting curve in \KARL, we first define a set of times $\mathcal{T}$ (zero to twenty days, in one-day increments). Then, for each time $t \in \mathcal{T}$, \KARL predicts the recall of a flashcard $f$ as if we were at time $t$, updating the features in $\mathcal{X}(f)$ accordingly.

\begin{figure}[t]
\centering
\includegraphics[width=\linewidth]{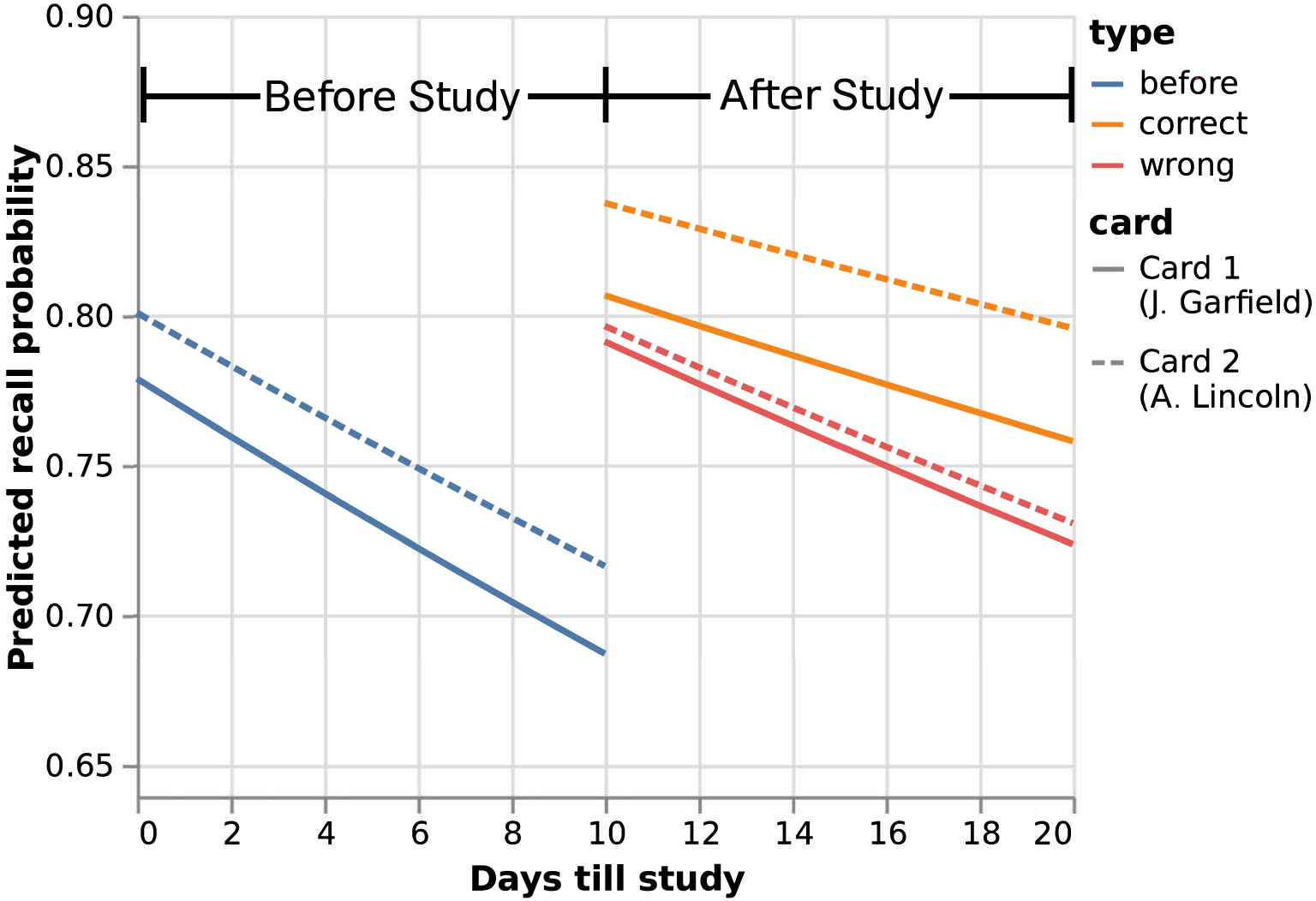}
\vspace{-3ex}
\caption{\label{fig:forgetting_curve}
Forgetting curve for US history cards. When Card 1 is studied, \KARL's prediction of the semantically related Card 2 increases, despite not being studied.
}
\vspace{-1ex}
\end{figure}

To see how BERT enables \KARL to make semantic inferences across flashcards, we show simulated forgetting curves for two related flashcards. In Figure~\ref{fig:forgetting_curve}, Card 1 (James Garfield, 20th U.S. president) is studied once on day 0 and again on day 10. When Card 1 is recalled correctly on~day 10, the recall prediction of Card 2 (Abraham Lincoln, 16th U.S. president) increases despite not being studied. This~highlights the benefit of making semantic inferences across flashcards: \KARL intelligently adjusts recall predictions across semantically-related flashcards based on the learner's study of just one.

\subsection{\KARL Retrieval Case Study} \label{subsection:qualitative_analysis2}

\begin{figure}[t]
\centering
\fbox{\includegraphics[width=0.96\linewidth]{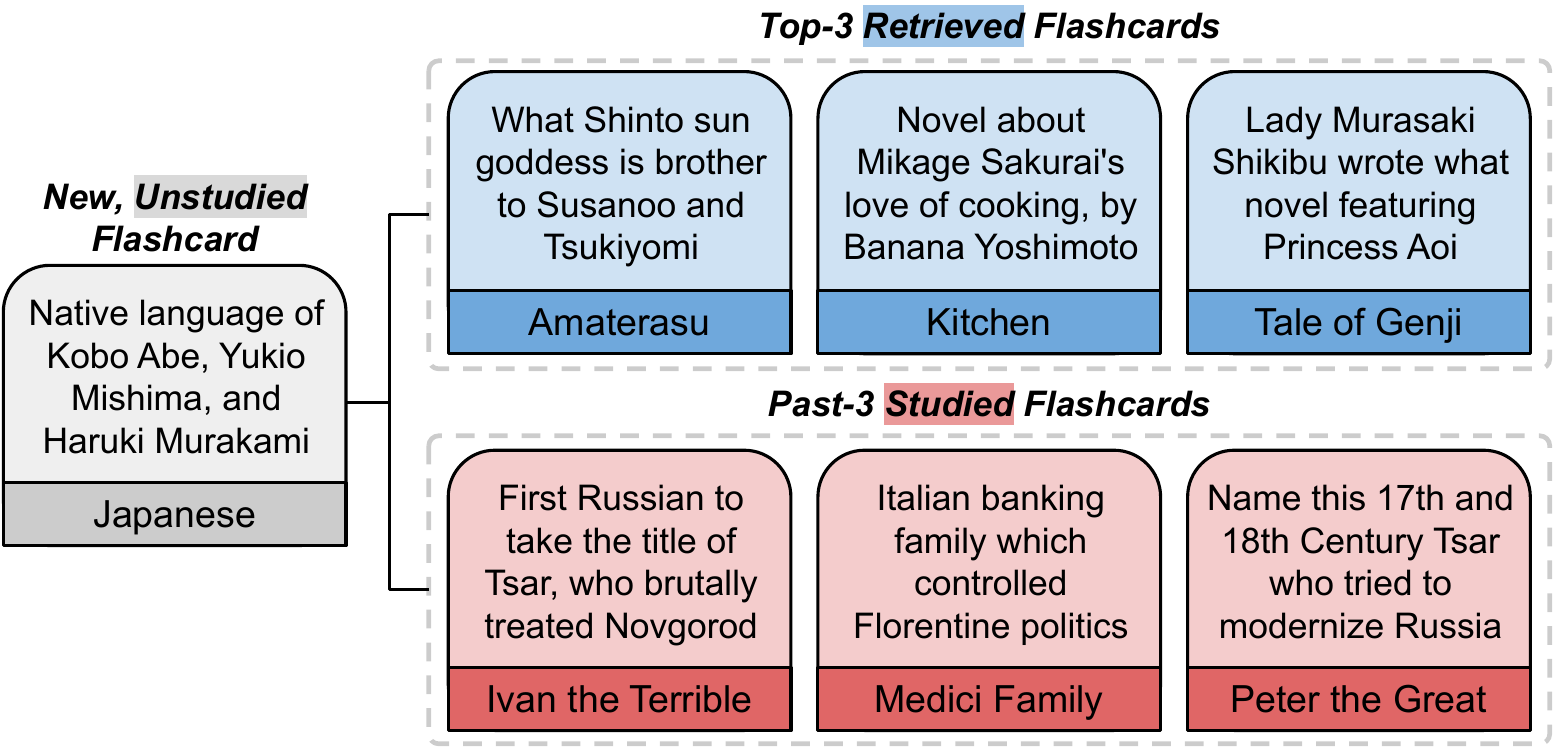}}
\vspace{-3.5ex}
\caption{\label{fig:dkt_concepts}
Top-3 retrieved vs past-3 studied flashcards when the user studies a new card on Japanese literature.
}
\vspace{-1ex}
\end{figure}

When using a subset of study history, some work uses the past-$k$ cards instead of retrieving $k$ cards \cite{pavlik2021logistic}.
Using the past-$k$ cards lowers AUC (Appendix~\ref{appendix:ablation_pastk}), but the following case study where the user sees a new card on Japanese authors also reveals the strength of retrieval (Figure~\ref{fig:dkt_concepts}).
The top-3 cards retrieved by \KARL are topically related to the current card (Japanese novels and Shinto), but the past-3 cards seen by the user are on European history.
Despite the topic shift, \KARL can still predict student recall, as the retrieved cards' study data reveal the student's knowledge of Japanese literature.
As \KARL groups similar cards with retrieval, \KARL can also propose new concepts in the current flashcard that the student may know, like a concept of Japanese culture. Thus, \KARL recovers concepts without the content annotations used in DKT datasets \cite{Liu2021ASO}, further showing the benefits of retrieval.

\section{Online Evaluation} \label{section:online_evaluation} 

Our offline evaluation (\cref{section:offline_evaluation}) shows \KARL predicts recall accurately, but this does not capture the main goal of any educational tool: \textbf{enhancing learning}.
Thus, while researchers often stop at offline evaluation to claim scheduler superiority, we now provide evidence that content-aware schedulers can improve student learning over FSRS, the SOTA~scheduler. 
We propose a teaching policy that equips DKT student models like \KARL for scheduling (\cref{subsection:teaching_policy}), design a test mode user study in our app to measure two facets of medium-term learning (\cref{subsection:test_mode_setup}), and assess student learning outcomes to compare the learning benefits of \KARL versus FSRS (\cref{subsection:test_mode_results}).

\subsection{A Delta-Based Teaching Policy for \KARL} \label{subsection:teaching_policy}

After \KARL predicts if the student can recall a given flashcard, a teaching policy decides \emph{when} to show this card next~\cite{hunzikerTeachingMultipleConcepts2019}. DKT models directly predict recall, so the only existing compatible policy for DKT models like \KARL is \textbf{threshold-based}, which picks flashcards near specified retention levels like 0.9~\cite{FSRS} or 0.94~\cite{eglingtonOptimizingPracticeScheduling2020}. A threshold-based policy for \KARL could run the classifier $p(a_t \g f_t, \mathcal{F}')$ on every flashcard and pick those with predicted recall near a specified retention level.
But without extra heuristics, this method will not schedule flashcards that \KARL predicts a student cannot recall or forgot (i.e. predicted recall of 0), even if studying such cards would aid subject mastery.

To solve this, we design a \textbf{delta-based teaching policy} that calculates how a student's recall improves over a given time interval $\Delta$ after studying a flashcard.
Concretely, given a candidate flashcard $f$ at time $t$, the policy uses \KARL to simulate~what the student's recall will be after $\Delta$ time, $t' = t + \Delta$, based on if the student does or does not study $f$ at~$t$.
By selecting cards with the highest differences in the recall predictions at $t'$, which can include cards with initially low predicted recall, the policy aligns with our core goal of maximizing student recall.

When \KARL predicts a user can recall $f$ correctly at time $t$ with probability $p_t(f)$, we aim to compute by how much their future recall $p_{t'}(f)$~will increase if $f$ is studied at $t$, called the \textbf{delta score:} \begin{equation}
\label{eq:learn_delta}
p_{t'}(f \mid \text{study } f \text{ at } t) - p_{t'}(f \mid \text{no study at } t).
\end{equation} To get ${p_{t'}(f \mid \text{no study } f \text{ at } t)}$, future recall probability when the user does not study, we obtain a prediction from \KARL as if we were at time $t'$.

For ${p_{t'}(f \mid \text{study } f \text{ at } t)}$, future recall when the user studies, we consider two possible outcomes of studying the card $f$ at time $t$: the student's answer could be correct or incorrect. We weigh these two outcomes by the model's initial prediction $p_t(f)$:\begin{equation}
\begin{aligned}
p_{t'}(f \mid &\text{ study } f \text{ at } t) \coloneqq \\
&p_{t'}(f \mid \text{correct at } t) \cdot p_t(f)\\
+\; &p_{t'}(f \mid \text{incorrect at } t) \cdot (1 - p_t(f)).
\end{aligned}
\end{equation} To find the two $p_{t'}$ values, we update the study data $\mathcal{X}(f)$ to query \KARL as~if we are at time $t'$ and if the student answered correctly/incorrectly at time~$t$.

The cards with the $n$-highest delta scores (Eq.~\ref{eq:learn_delta}) are the next $n$ cards scheduled. With a delta-based teaching policy, we ensure flashcards are picked~for maximizing learning, preventing cards from~being neglected solely due to having low predicted recall.

\subsection{Test Mode Setup} \label{subsection:test_mode_setup}

We deploy FSRS and \KARL with our delta teaching policy (i.e. \KARLDelta), with $\Delta$ equal to 1 day. We recruit users with the same procedure as \cref{subsection:collecting_student_data} and award all users who complete our study \$50.  Studying how schedulers impact learning is difficult, as: (1) users want to study diverse subjects; (2) users have varying background knowledge; and (3) learning, a multifaceted process, is hard to quantify.

For (1), we create two test sets fixed across users, both with 20 manually-written cards from seven diverse QANTA bonus questions \cite{qblink}; bonus questions in QANTA are grouped as three distinct subquestions that test the same concepts, forming a testbed for evaluating the ability of content-aware schedulers to make semantic inferences. Users are tested on the same cards, which likely have some facts of interest to the user. For (2), we use a within-subject design \cite{lindsey2014improving}, where users study with \KARL for one test set, and FSRS for the other. Hence, users~try both schedulers, preserving background knowledge.

For (3), we focus on \textbf{medium-term learning}, which we define as the ability to learn flashcards over several days. 
This goal reflects the time span students report preparing for exams; 
students often study flashcards for five days or less and study no more than half their flashcards at a time \cite{wissman2012and}.
Mirroring this, our users review ten cards from the test set of 20, scheduled by FSRS or \KARLDelta, daily for five days.
On day six, users complete a post-test and study all 20 test set cards.

Medium-term learning has not been deeply studied, so we use two measures of memory strength that have been explored in prior research: response accuracy and response time \cite{macleod1984response, maris2012speed}.
We define \textbf{response accuracy} \cite{alshammari2019design} as the gain in accuracy on the post-test from when users first see cards (pre-test).
Higher response accuracy means that the user has learned more flashcards.

Response accuracy captures \textit{how many} facts a user has learned, but not \textit{how well} or \textit{quickly} they can recall these facts.
Thus, we define \textbf{response time} as the mean time needed for users to recall flashcard answers \cite{ericsson1985memory}---the time from when the card is viewed until when the answer is submitted. 
We calculate response time on two splits: flashcards answered correctly; and all flashcards studied.
Lower response time on cards answered \emph{correctly} means the user is more familiar with cards they know, since they recall correct answers faster, while lower response time on \emph{all} cards means that the user completes the post-test faster.

\begin{table}[t]
\small
\centering
\setlength{\tabcolsep}{3pt}
\renewcommand{\arraystretch}{0.8}
\begin{tabular}{@{}ccccc@{}}
\toprule
\textbf{Model} & \textbf{Pre/Post-Test Acc} & \textbf{Response Corr/All} & \textbf{TTP} \\ \midrule
FSRS            & 0.41 / \textbf{0.88} & 6.58 sec / 6.82 sec  & 2.58       \\
\KARL+$\Delta$  & \textbf{0.42} / 0.86 &  \textbf{6.15 sec} / \textbf{6.27* sec} & \textbf{2.74}        \\ \bottomrule
\end{tabular}
\vspace{-2ex}
\caption{Post-test metrics for \KARLDelta vs FSRS users. Best in \textbf{bold}. * means $t$-test significance ($p < 0.05$). \KARLDelta users have similar pre-test and post-test accuracy versus FSRS users but much lower post-test response time, enhancing testing throughput (TTP).}
\label{table:phase_2}
\vspace{-0.5ex}
\end{table}


We combine response accuracy and time into a single post-test performance metric \cite{chignell2014combining} and define \textbf{testing throughput (TTP)} as the average total correct answers ($20\;*$ post-test accuracy) divided by the average response time on all cards---cards answered correctly per second spent on the post-test. Higher TTP indicates that the student has a deeper understanding of the test mode flashcards, as they can correctly answer the same number of flashcards while taking less time.

\subsection{User Study Results} \label{subsection:test_mode_results}

We collect 32 six-day study sessions from 27 students with FSRS and \KARLDelta (Table~\ref{table:phase_2}). Both schedulers help users achieve similar mastery in test mode, more than doubling pre-test to post-test accuracy ($0.42$ to $0.86$). While response accuracy is similar, \KARLDelta has lower post-test response time on cards answered correctly, suggesting that \KARL users are more familiar with the cards answered correctly. Finally, \KARLDelta users maintain post-test accuracy while reducing recall time on all cards studied. As a result, \KARLDelta users obtain higher testing throughput (TTP) than FSRS users, as users studying with \KARL answer a similar number of facts  correctly in less time ($2.74$ vs $2.58$ flashcards answered correctly per second). 

Combining offline (Table~\ref{table:quant_results}) and online (Table~\ref{table:phase_2}) results, \KARLDelta: 1) produces more accurate and calibrated recall predictions than FSRS on seen cards; 2) can predict recall on unseen cards, unlike FSRS; 3)~matches FSRS in enhancing response accuracy; and 4) enables users to recall answers faster than FSRS, increasing testing throughput. Holistically, \KARLDelta bests FSRS. 
Since \KARLDelta is a baseline content-aware scheduler and it already rivals SOTA, content-aware scheduling is a promising~paradigm, which we hope will motivate works to build better content-aware schedulers and look beyond study data to fully capture student abilities.


\section{Conclusion}


We introduce and successfully implement the first content-aware scheduler with \KARLDelta, a simple but effective model using DKT, BERT, retrieval, and a novel delta-based teaching policy.
Our offline evaluation on a newly-collected dataset shows \KARL provides accurate and well-calibrated recall predictions, while our online evaluation reveals \KARL improves testing throughput over SOTA.
Thus, we give the first evidence that content-aware schedulers can be used to improve student learning.

Content-aware scheduling enhances personalization in learning tools, as models can infer student knowledge gaps through semantic inferences.
Given this strength, we hope future works extend content to modalities beyond text, such as images or audio.
In online evaluation, \KARL surpasses FSRS in testing throughput, but there are still many facets of learning that can be measured with online studies; these learning metrics could not only be used for evaluation but also training signals.
In all, we show the strength of content-aware scheduling, which we hope will motivate works to look beyond \textit{single} study items and also model relations \textit{between} study items to fully capture student~abilities.


\section{Limitations}

One limitation of \KARL, which is shared with all DKT models, is that our model uses BERT representations and a neural classifier, resulting in a slower inference time compared to student models which only use flashcard-level features. To minimize this limitation, we consider two design choices. First, rather than embedding the entire student history like existing DKT models, we perform top-$k$ retrieval, enabling \KARL to have a consistent inference time that does not scale with the size of the student's study history. Second, we implement our retriever operations with FAISS \cite{johnson2019billion}, an efficient vector database. This allows us to quickly look up the representation of each flashcard derived from QANTA, eliminating the time needed to tokenize and feed each flashcard through BERT, and efficiently perform Maximum Inner-Product Search. During our test mode user study, we did not receive complaints about the inference time or efficiency of \KARL.

Further, while \KARL is an effective framework for student modeling on our diverse dataset on trivia questions, we have not analyzed the performance of \KARL on domain-specific DKT benchmarks that do not have easily accessible textual content, such as EdNet \cite{choi2020ednet} or ASSISTments\footnote{To obtain the textual content of ASSIStments, you must send an additional email for verification and agree to a Terms of Use before your request can even be processed. In contrast, our dataset will release the flashcard content for public use.} \cite{selent2016assistments}. Since the primary goal of this work was to design a general-purpose content-aware scheduler, we require a diverse and content-focused dataset to evaluate our model. In future works, when more datasets provide textual content, it would be interesting to study if the accuracy and calibration of \KARL is still strong, or how transfer learning techniques \cite{weiss2016survey} can help \KARL adapt to specific domains.

Finally, exploring more advanced retrievers and language models could improve the predictions of \KARL even further. As this was the first demonstration of content-aware scheduling and using retrieval for flashcard learning, we focused on the most basic language model and retriever: an off-the-shelf BERT model. We consider it a positive sign that this simple design leads to large offline AUC and ECE gains, which will motivate future works, including new iterations of \KARL, that improve content-aware scheduling with more advanced retrieval techniques and language models.

\section{Ethical Considerations}

The goal of adaptive student models like \KARL is to make personalized predictions about a student's knowledge level. When making such tailored predictions, it is crucial to ensure that these student models are not exploiting private information about the user. To ensure that models trained on our dataset do not succumb to this risk, our dataset only identifies users by a numerical ID. Further, our data collection and test mode user study were both approved by an Institutional Review Board (IRB), allowing us to fully address any potential risks of our study. All users were compensated based on a raffle system, and could win either \$10, \$15, or \$50 based on the study. In our advertisements, it was made clear to users before signing up that they would be part of a research study. 

\section*{Acknowledgements}

We would like to thank members of the CLIP lab at the University of Maryland and external collaborators for their discussions of this work, including Jarrett Ye, Jack Wang, David Martinez, Dayeon Ki, Yu Hou, Yoonjoo Lee, Hyojung Han, Haozhe An, Erik Harpstead, Paulo Carvalho, Rachel Rudinger, Pedro Rodriguez, Denis Peskoff, Roger Craig, Sydney Montoya, and Wenyan Li. We additionally thank Nathaniel Klein for helping name \KARL.
This material is based upon work supported by the National Science
Foundation under Grant No. \abr{iis}-2403436 (Boyd-Graber) and \abr{dge}-2236417 (Balepur).
Any opinions, findings, and conclusions or recommendations expressed
in this material are those of the author(s) and do not necessarily
reflect the views of the National Science Foundation.
Cloud computing resources were made possible by a gift from Adobe Research.
%

\bibliography{custom}\
\bibliographystyle{acl_natbib}

\appendix

\clearpage

\section{Appendix}

\subsection{Flashcard Creation and Examples} \label{appendix:subsection:flashcard_examples}

Each question $q$ in QANTA consists of a list of clues $q = \{c_1, ..., c_n\}$ describing a single answer $\mathcal{A}$. To convert $q$ into a flashcard $f$, we first map each unique $\mathcal{A}$ to its corresponding questions $\mathcal{Q} = \{q_1, ..., q_m\}$. Next, we take a random sample of clues $\mathcal{C} = \{c_1, ..., c_p\}$ from $\mathcal{Q}$. Using this data, we create $f$, where the clue $c \in \mathcal{C}$ is the front of $f$ (i.e. the question shown to the user) and the answer $\mathcal{A}$ is the back of $f$. Repeating this process for the entire dataset, we obtain 23,918 unique flashcards spanning 11 diverse subjects.

In Table~\ref{appendix:table:flashcards}, we provide examples of the flashcards created from the QANTA dataset spanning 11 unique topics. The QANTA dataset is publicly available and we used the dataset with its intended research use. These flashcards are deployed into \KARL, where 543 participants produced 123,143 study records. In these records, 44.02 percent contain studies on new flashcards and students correctly answer the flashcard shown 71.80 percent of the time. On average, each user studied 226.78 flashcards during the data collection period.

The data used to create our flashcards is well-established and was not significantly altered in this work, so we did not check if the dataset uniquely identifies individuals or contains offensive content. All of our flashcards are written in English, containing facts primarily targeted toward students in the American high school and college education systems. The only additional data we collected beyond this that is released is the user's ID, the date and time of study, and which flashcards this user got correct and incorrect, which pose no harms. 

In Tables~\ref{appendix:table:data_qual}, \ref{appendix:table:data_quant}, and \ref{appendix:table:data_time} we show the descriptions and summary statistics of the qualitative, quantitative, and time-based feature columns in our released dataset, respectively. 

\subsection{\KARL Classifier Features} \label{appendix:subsection:classifier_features}

Our released dataset also includes all the hand-picked features that \KARL uses along with BERT representations (Table \ref{appendix:table:data_quant}). We normalize each feature using mean value and standard deviation computed on the training set. Features with IDs 7.17 through 7.22 are computed using the Leitner \cite{leitner1974so} and SM-2 \cite{wozniak1990optimization} algorithms. 

In each study session, \KARLDelta additionally tracks the session-level variants of features 7.11-13. 
In test mode, features such as the user's accuracy are frozen to ensure that \KARLDelta does not have an unfair advantage compared to FSRS.

For all offline experiments except for the ablation, \KARL is trained on features 7.1, 7.4, 7.7, 7.10-14, and 7.16. For the ablation and online evaluation, \KARL is trained on all the features.

\subsection{DKT Model for User Study} \label{appendix:cold_start}

One of the models we in our user study to collect our flashcard dataset was a DKT model trained on the Protobowl dataset \cite{Rodriguez2019QuizbowlTC}.
The Protobowl dataset contains QANTA questions with human responses to these questions, but this can be adapted to a flashcard dataset for content-aware models by selecting instances where users answered and practiced with the same question, treating these questions as study items.
After converting the dataset to this format, we trained a DKT model similar to \KARL, but it does not use retrieval, and thus only uses BERT and flashcard features on the current flashcard to predict recall.

We designed this model to study whether it was possible to design a content-aware scheduler by only adapting existing datasets, rather than collecting a new one.
While we found this to be feasible, training directly on a specialized flashcard dataset instead of using some dataset adaptation techniques would expectedly improve the performance of a content-aware flashcard scheduler.
Thus, we opted to collect a new dataset and train a content-aware model (\KARL) on this dataset for the best possible results in the online user study.

\subsection{Forgetting Curves} \label{appendix:subsection:forgetting_curves}

\begin{figure*}
    \centering
    \includegraphics[width=\linewidth]{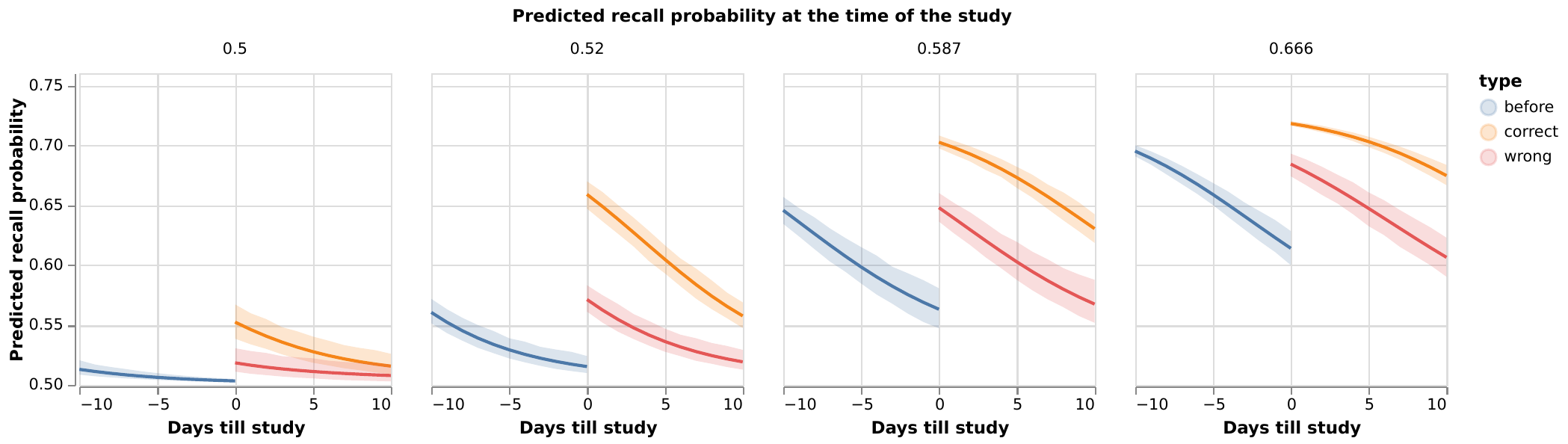}
    \caption{The average forgetting curve ten days before and after a study (day zero) for both when the user succeeds and fails at recalling the flashcard. Unlike exponential forgetting models, the convexity of our forgetting curve depends on both the current predicted recall and the outcome of the most recent study, adding more flexibility.}
    \label{appendix:figure:more_forgetting_curves}
\end{figure*}

In this section, we provide more examples of forgetting curves. Unlike previous work which forces the student model to follow a specific forgetting function, we consider student modeling as predicting the binary outcome of each study. In Figure~\ref{appendix:figure:more_forgetting_curves}, we showcase the flexibility of our forgetting curves, and display visualizations of how the forgetting curve of a flashcard changes depending on its predicted recall probability.

\subsection{Past-$k$ Ablations}
\label{appendix:ablation_pastk}

In Table~\ref{appendix:table:ablation_pastk}, we present an additional ablation where we use top-$k$ retrieval versus past-$k$ retrieval in \KARL.
We find that using top-$k$ retrieval outperforms past-$k$ retrieval on 3/4 metrics, suggesting that retrieval returns more relevant study items than just using the past-$k$ items that the user has studied.
Since the gap between these two methods is relatively small, we intuit that retrieval only shows clearer advantages over past-$k$ retrieval when a user shifts topics while studying, like in the case study in \cref{subsection:qualitative_analysis2}.
In fact, we find that only $5.7\%$ of our dataset has topic shifts; this suggests that topic shifts are relatively rare, explaining why the performance gains are small.
However, retrieval still boosts performance, so we argue that empirically it is an effective design choice.



\subsection{Embedding Model Comparison}
\label{appendix:ablation_llama}

In Table~\ref{appendix:table:ablation_llama}, we compare the original \KARL BERT model, with and without top-5 retrieval, with a \KARL model using LLaMA embeddings.
LLaMA returns an embedding of size 5192, so we were unable to train a version with retrieval using these embeddings due to resource constraints.
We find that on seen cards, the AUC and ECE of both BERT variants outperforms LLaMA; this suggests that the model may be overweighting LLaMA embedding importance compared to the student's study data.
While the BERT model with retrieval performs better on unseen cards, its performance on seen cards is similar to the model without retrieval. This aligns with out intuition that retrieval benefits unseen card prediction while past study data for seen cards is sufficient.
However, on unseen cards, KARL LLaMA without retrieval has stronger than AUC than KARL BERT with retrieval, showing that models that can make stronger semantic inferences will have more accurate predictions on unseen cards. 
Note though that LLaMA has worse ECE, suggesting that larger LLMs are more overconfident when predicting student recall.
This confirm the strength of our retrieval-augmented method, as gains cannot be achieved just by scaling the embedding model without retrieval.

\subsection{Retriever Method Comparison}
\label{appendix:ablation_retriever}

In Table~\ref{appendix:table:ablation_retriever}, we compare \KARL's BERT semantic similarity retrieval method with an alternative BM-25 retriever. We find that the original model using BERT similarity outperforms BM-25 across all metrics, confirming the notion that more advanced retrievers are better at identifying relevant items in a student's study history.

\subsection{Training Details} \label{appendix:subsection:training}

We train \KARL for 12 hours using a single NVIDIA RTX:A4000 GPU. Some model variants were trained using a single NVIDIA A100 GPU with 80 GB GPU memory. Parameters were manually selected without search. We use the default BERT configuration, and our classifier is implemented in PyTorch\footnote{\url{https://pytorch.org/}} with the following layers: 1) BERT model; 2) Dropout; 3) Linear Layer; 4) GELU Activation Layer; 5) Layer Normalization; 6) Dropout; 7) Linear Layer. We minimize the binary cross-entropy loss of this model. We use the Adam optimizer \cite{KingBa15}, a learning rate of $0.00005$, a batch size of 64,and 10 epochs.

All baselines are implemented using the official code provided by the authors of the respective papers, and all hyperparameters in the model implementations were chosen according to the reported hyperparameters in the paper. LM-KT was trained on a single NVIDIA RTX:A6000 GPU and training parameters were the default values in the provided code. FSRS was trained using the Collaboratory notebook provided by the authors.

All metrics were reported from a single run. Accuracy, AUROC, and Expected Calibration Error were all computed using scikit-learn.\footnote{\url{https://scikit-learn.org/}}

\subsection{User Study Instructions}

\begin{figure*}
    \centering
    \fbox{\includegraphics[width=\linewidth]{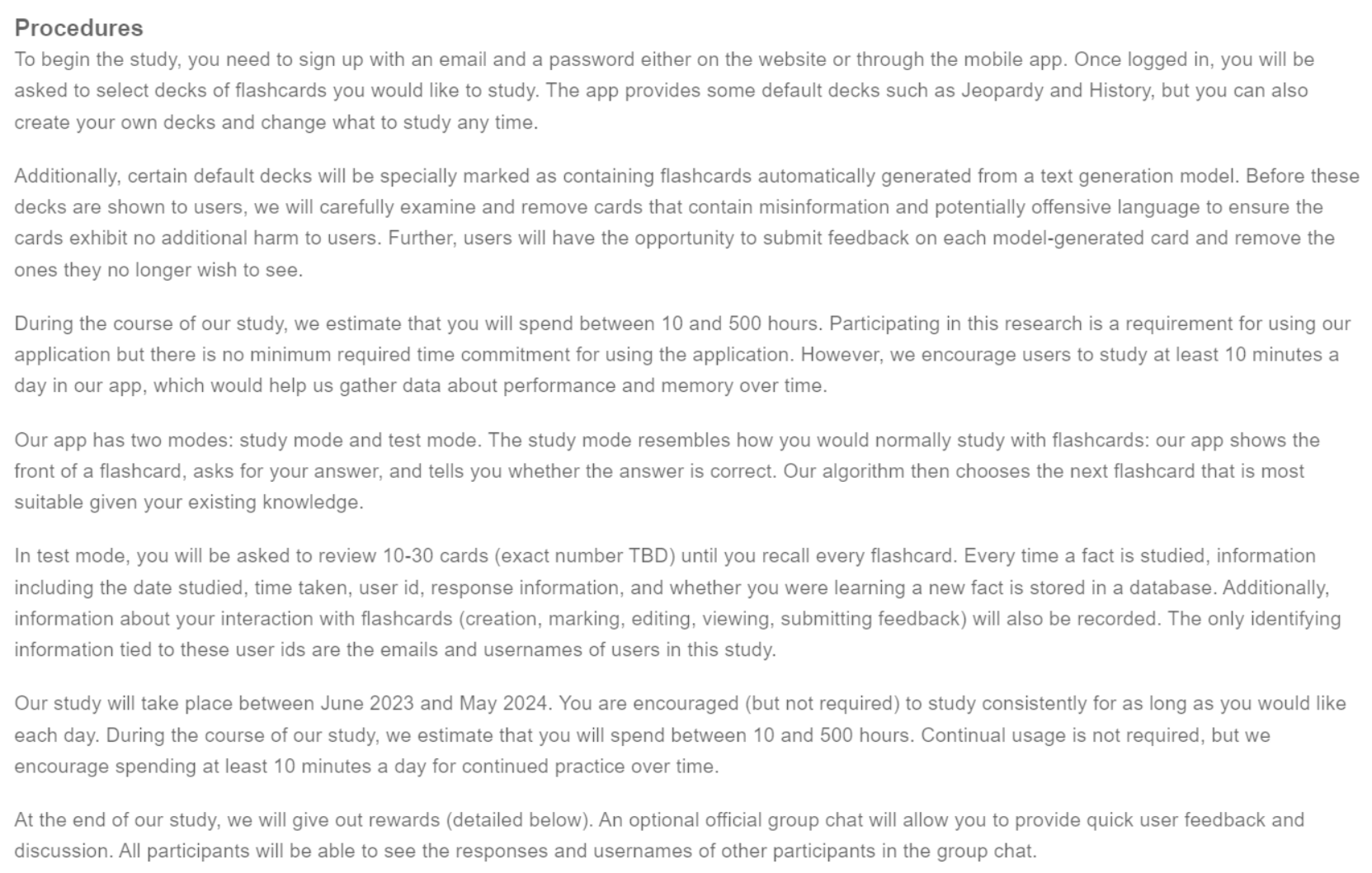}}
    \caption{Procedures and instructions shown to users.}
    \label{appendix:figure:IRB}
\end{figure*}

\begin{figure*}
    \centering
    \fbox{\includegraphics[width=\linewidth]{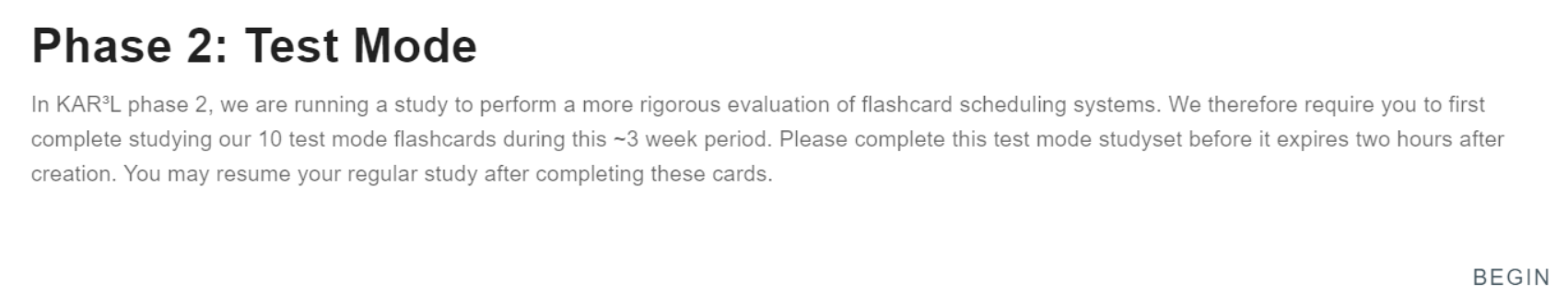}}
    \caption{Test Mode popup instructions shown to users just before completing a test mode study.}
    \label{appendix:figure:phase2}
\end{figure*}

\begin{figure*}
    \centering
    \fbox{\includegraphics[width=\linewidth]{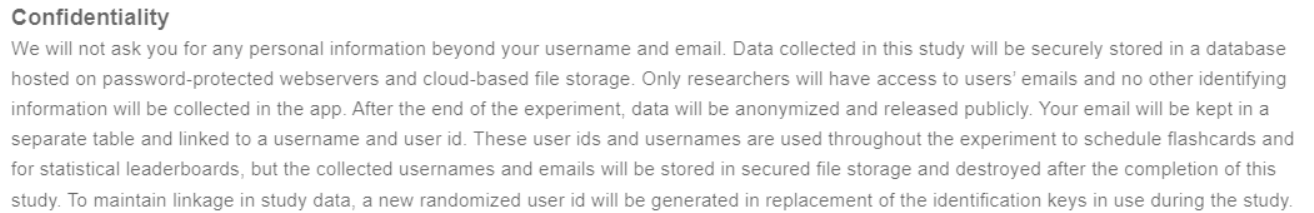}}
    \caption{Confidentiality details shown to users.}
    \label{appendix:figure:confidentiality}
\end{figure*}

During our user studies, we ensure to provide clear instructions to our users. First, on the home page of our app, the users can read the procedures of our IRB (Figure~\ref{appendix:figure:IRB}). This serves as detailed instructions for our user studies. Second, when it is time for the user to study test mode, a popup appears giving brief instructions about the next test set they are about to study (Figure~\ref{appendix:figure:phase2}). Users were also made aware of the confidentiality of their data (Figure~\ref{appendix:figure:confidentiality}), which can be viewed by clicking on the IRB link on our home page. The rest of the instructions are outlined in our advertisements.

\clearpage
\begin{table*}[t]
\centering
\resizebox{\textwidth}{!}{%
    \begin{tabular}{l l l c}
     \toprule
     \textbf{Category} & \textbf{Example Front (Question)} & \textbf{Example Back (Answer)} & \textbf{Category Frequency} \\ \midrule
     Philosophy & \specialcellleft{William James and, later, Richard Rorty continued \\ a strain of philosophy largely inaugurated by this philosopher} & C.S. Peirce & 425  \\ \midrule
     Trash (Pop Culture) & \specialcellleft{Recently-cancelled FOX show which starred Eliza Dushku \\ as Echo and which was directed by Joss Whedon} & Dollhouse & 496 \\ \midrule
     Mythology & \specialcellleft{To pay Thrym back for stealing Mjolnir, Thor wore bridal \\ clothes to disguise himself as this goddess} & Freya & 1194 \\ \midrule
     Science & \specialcellleft{Metal which is used in aerospace applications due to its \\ durability, with atomic number 22 and symbol Ti} & Titanium & 2423 \\ \midrule
     Current Events & \specialcellleft{A citizen of this nation crashed a plane into the \\ French Alps in March 2015} & Federal Republic of Germany & 99 \\ \midrule
     Fine Arts & \specialcellleft{Revived by Wanda Landowski, it helped bring success \\ to its 1955 performer, Glenn Gould} & the Goldberg Variartions & 3158 \\ \midrule
     Religion & \specialcellleft{This text was originally written in unknown characters \\ referred to as "Reformed Egyptian"} & The Book of Mormon & 807 \\ \midrule
     Literature & \specialcellleft{Chief harpooner of the Pequod, a cannibal companion of \\ Ishmael in Melville's Moby Dick} & Queequeg & 4822 \\ \midrule
     Social Science & \specialcellleft{Luigi Pasinetti created a fifteen-equation mathematical \\ model of this economist's views} & David Ricardo & 966 \\ \midrule
     Georgraphy & \specialcellleft{The Franz Joseph and Fox Glaciers can be found in this nation,\\ notable for existing at low altitudes} & New Zealand & 1053 \\ \midrule
     History & \specialcellleft{This figure ruled during the Interregnum and led the New Model Army \\ after the assassination of Charles I} & Oliver Cromwell & 8476 \\ \bottomrule
    \end{tabular}
}
\caption{Examples of flashcards spanning 11 topics derived from the QANTA dataset.}
\label{appendix:table:flashcards}
\end{table*}
\begin{table*}[]
\footnotesize
\centering
\begin{tabular}{@{}c|l|l|c@{}}
\toprule
\textbf{ID} & \multicolumn{1}{c|}{\textbf{Column}} & \multicolumn{1}{c|}{\textbf{Description}} & \multicolumn{1}{c}{\textbf{Num Unique}} \\ \midrule
6.1 & user\_id & Identifier for the user & 543 \\
6.2 & card\_id & Identifier for the flashcard & 18663 \\
6.3 & card\_text & Text on the flashcard & 17554 \\
6.4 & deck\_id & Identifier for the deck (subject) being studied & 71 \\
6.5 & deck\_name & Name of the deck (subject) being studied & 60 \\ \bottomrule
\end{tabular}
\caption{Qualitative value columns in our released dataset, along with descriptions and number of unique values.}
\label{appendix:table:data_qual}
\end{table*}

\begin{table*}[]
\footnotesize
\centering
\setlength{\tabcolsep}{2.5pt}
\begin{tabular}{@{}c|l|l|ccc@{}}
\toprule
\multicolumn{1}{c|}{\textbf{ID}} & \multicolumn{1}{c|}{\textbf{Column}} & \multicolumn{1}{c|}{\textbf{Description}} & \textbf{Mean} & \textbf{Min} & \textbf{Max} \\ \midrule
7.1 & is\_new\_fact                        & Whether the user has (not) seen this card before                                                                         & 0.44             & 0.00         & 1.00         \\
7.2 & user\_n\_study\_positive             & Number of times the user has studied and answered correctly                                      & 1.95e03          & 0.00         & 1.47e04     \\
7.3 & user\_n\_study\_negative             & Number of times the user has studied and answered incorrectly                                    & 471.14           & 0.00         & 3.52e03      \\
7.4 & user\_n\_study\_total                & Number of times the user has studied in total                                                              & 2.42+e03          & 0.00         & 1.63e04     \\
7.5 & card\_n\_study\_positive             & Number of times the card has been answered correctly                                                   & 6.78             & 0.00         & 82.00        \\
7.6 & card\_n\_study\_negative             & Number of times the card has been answered incorrectly                                                & 1.93             & 0.00         & 35.00        \\
7.7 & card\_n\_study\_total                & Number of times the card has been studied in total                                              & 8.72             & 0.00         & 92.00        \\

7.8 & usercard\_n\_study\_positive             & Number of times the user answered this card correctly                                                   & 1.04             & 0.00         & 24.00        \\
7.9 & usercard\_n\_study\_negative             & Number of times the user answered this card incorrectly                                                & 0.60             & 0.00         & 19.00        \\
7.10 & usercard\_n\_study\_total                & Number of times the user answered this card in total                                              & 1.64            & 0.00         & 26.00        \\

7.11 & acc\_user                            & Accuracy of this user in all studies                                                                  & 0.70             & 0.00         & 1.00         \\
7.12 & acc\_card                            & Accuracy of all users on this card                                                                     & 0.60             & 0.00         & 1.00         \\
7.13 & acc\_usercard                        & Accuracy of the user on this card                                                                                       & 0.33             & 0.00         & 1.00         \\
7.14 & usercard\_delta                      & Time (hours) since the previous study of this card by this user. & 194.85           & 0.00         & 6.79e03      \\
7.15 & usercard\_delta\_previous            & The previous delta. Zero if first or second study.                                                                       & 2.92e05         & 0.00         & 1.52e07     \\
7.16 & usercard\_prev\_response  & The result (correct/incorrect) of the previous study                                                                     & 0.37             & 0.00         & 1.00         \\

7.17 & leitner\_box             & Partition the user is in according to the Leitner system                                                   & 1.48             & 0.00         & 10.00        \\
7.18 & sm2\_efactor             & Easiness factor computed by SM-2                                                 & 1.15             & 0.00         & 2.50        \\
7.19 & sm2\_interval                & Interval computed by SM-2                                              & 1.26e05            & 0.00         & 1.86e08        \\
7.20 & sm2\_repetition                & Repetition factor computed by SM-2                                              & 0.88            & 0.00         & 24.00        \\
7.21 & delta\_to\_leitner             & Days until Leitner would schedule the card for review                                                & 64.21             & -6.8e03         & 2.40e04        \\
7.22 & delta\_to\_sm2                & Days until SM-2 would schedule the card for review                                              & 2.91e05            & -97.00         & 4.58e07        \\

7.23 & elapsed\_milliseconds                & Time user spent thinking about the answer before submitting    & 1.04e04         & 0.00         & 6.00+e04     \\
7.24 & n\_minutes\_spent                & Number of minutes the user spent on the app    & 353.28         & 0.00         & 2.37e03     \\
7.25 & correct\_on\_first\_try              & Did the user answer the card correctly on the very first study.                                                     & 0.26             & 0.00         & 1.00         \\
7.26 & response                             & Correct / Incorrect                                                                                                      & 0.71             & 0.00         & 1.00         \\ \bottomrule
\end{tabular}
\caption{Quantitative value columns in our released dataset, along with descriptions, min, max, and mean values.}
\label{appendix:table:data_quant}
\end{table*}
\begin{table*}[]
\footnotesize
\centering
\begin{tabular}{@{}c|l|l@{}}
\toprule
\textbf{ID} & \multicolumn{1}{c|}{\textbf{Column}} & \multicolumn{1}{c}{\textbf{Description}} \\ \midrule
8.1 & utc\_datetime & Datetime object for when the study of this card occurred \\
8.2 & utc\_date & Date version of utc\_datetime \\ \bottomrule
\end{tabular}
\caption{Timestamp data columns in our released dataset.}
\label{appendix:table:data_time}
\end{table*}

\begin{table*}[t]
\centering
\begin{tabular}{@{}lcccc@{}}
 & \multicolumn{2}{c}{\textit{Seen Cards}} & \multicolumn{2}{c}{\textit{Unseen Cards}} \\ \toprule
\multicolumn{1}{l|}{Model} & AUC ($\uparrow$) & \multicolumn{1}{c|}{ECE ($\downarrow$)} & AUC ($\uparrow$) & ECE ($\downarrow$) \\ \midrule
\multicolumn{1}{l|}{\KARL Full} & \textbf{0.780} & \multicolumn{1}{c|}{\textbf{0.108}} & \textbf{0.740} & 0.124 \\
\multicolumn{1}{l|}{Past-$k$} & 0.758 & \multicolumn{1}{c|}{0.119} & 0.729 & \textbf{0.119} \\
\multicolumn{1}{l|}{No BERT} & 0.692 & \multicolumn{1}{c|}{0.127} & 0.612 & 0.205 \\
\multicolumn{1}{l|}{No $\mathcal{X}(z)$} & 0.680 & \multicolumn{1}{c|}{0.135} & 0.620 & 0.191 \\ \bottomrule
\end{tabular}
\caption{\KARL ablations including past-$k$ retrieval versus top-$k$ (\KARL Full) retrieval. The No $\mathcal{X}(z)$ and No BERT ablations retain the top-$k$ retrieval. $k=5$ for all models.}
\label{appendix:table:ablation_pastk}
\end{table*}
\begin{table*}[t]
\centering
\begin{tabular}{@{}lcccc@{}}
 & \multicolumn{2}{c}{\textit{Seen Cards}} & \multicolumn{2}{c}{\textit{Unseen Cards}} \\ \toprule
\multicolumn{1}{l|}{Embedding Model} & AUC ($\uparrow$) & \multicolumn{1}{c|}{ECE ($\downarrow$)} & AUC ($\uparrow$) & ECE ($\downarrow$) \\ \midrule
\multicolumn{1}{l|}{BERT (w/ Retrieval)} & \textbf{0.864} & \multicolumn{1}{c|}{\textbf{0.091}} & 0.786 & \textbf{0.085} \\
\multicolumn{1}{l|}{BERT (no Retrieval)} & \textbf{0.864} & \multicolumn{1}{c|}{0.098} & 0.776 & 0.11 \\
\multicolumn{1}{l|}{LLaMA (no Retrieval)} & 0.841 & \multicolumn{1}{c|}{0.107} & \textbf{0.810} & 0.169 \\ \bottomrule
\end{tabular}
\caption{Comparison of \KARL models using BERT and LLaMA embeddings. A LLaMA model with retrieval could not be trained due to OOM error (limited GPU resources).}
\label{appendix:table:ablation_llama}
\end{table*}
\begin{table*}[t]
\centering
\begin{tabular}{@{}lcccc@{}}
 & \multicolumn{2}{c}{\textit{Seen Cards}} & \multicolumn{2}{c}{\textit{Unseen Cards}} \\ \toprule
\multicolumn{1}{l|}{Retrieval Method} & AUC ($\uparrow$) & \multicolumn{1}{c|}{ECE ($\downarrow$)} & AUC ($\uparrow$) & ECE ($\downarrow$) \\ \midrule
\multicolumn{1}{l|}{BERT} & \textbf{0.864} & \multicolumn{1}{c|}{\textbf{0.091}} & \textbf{0.786} & \textbf{0.085} \\
\multicolumn{1}{l|}{BM-25} & 0.858 & \multicolumn{1}{c|}{0.106} & 0.771 & 0.139 \\ \bottomrule
\end{tabular}
\caption{Comparison of \KARL using different dense and sparse retrieval methods: BERT and BM-25.}
\label{appendix:table:ablation_retriever}
\end{table*}

\end{document}